\documentclass[runningheads,12pt]{llncs}
\usepackage{xcolor}
\usepackage[T1]{fontenc}
\usepackage{graphicx}
\usepackage[utf8]{inputenc}
\usepackage[english]{babel}

\usepackage[text={16cm, 22.5cm}, centering]{geometry}

\usepackage{amsmath}
\usepackage{amssymb}
\usepackage{amsfonts}
\usepackage{graphicx}
\usepackage[colorlinks=true, allcolors=blue]{hyperref}

\newtheorem{Thm}{Theorem}

\newtheorem{Lemma}[Thm]{Lemma}
\newtheorem{Prop}[Thm]{Proposition}
\newtheorem{Cor}[Thm]{Corollary}

\newtheorem{defn}[Thm]{Definition}

\newenvironment{proofV}{\noindent{\it Proof.}}{\qed\vspace{5pt}}
\newenvironment{Def}{\begin{defn}\normalfont}{\hfill$\dashv$\end{defn}}
\newenvironment{Remark}{\begin{remark}\normalfont}{\hfill$\dashv$\end{remark}}
\newenvironment{Example}{\begin{example}\normalfont}{\hfill$\dashv$\end{example}}

\usepackage{enumitem}

\newcommand{\f}{\varphi}
\newcommand{\la}{\langle}
\newcommand{\ra}{\rangle}

\newcommand{\Z}{\mathbb{Z}} 
\newcommand{\N}{\mathbb{N}} 
\newcommand{\FF}{\mathcal{F}} 
\renewcommand{\SS}{\mathcal{S}} 

\newcommand{\XX}{\mathcal{X}} 
\newcommand{\rest}{\!\restriction\!}
\newcommand{\cat}{{}^\frown}

\newcommand{\II}{\mathcal{I}}

\newcommand{\es}{\varnothing}

\newcommand{\dom}{\operatorname{dom}}
\newcommand{\MSR}{\operatorname{MSR}}
\newcommand{\id}{\operatorname{id}}
\renewcommand{\le}{\leqslant}

\usepackage{subcaption}
\allowdisplaybreaks

\begin{document}
\title{An Internal Model Principle For Robots}
%
%
\author{Vadim K. Weinstein, 
Tamara Alshammari, 
Kalle G. Timperi, \\
Mehdi Bennis, 
Steven M. LaValle
}
\authorrunning{V. K. Weinstein et al.}
%
\institute{University of Oulu} 
%
\maketitle              
\begin{abstract}
  When designing a robot's internal system, one
  often makes assumptions about the structure of the intended 
  environment of the robot. One may even assign 
  meaning to various internal components of the robot in terms
  of expected environmental correlates. 
  In this paper we want to make the distinction between
  robot's internal and external worlds clear-cut. 
  Can the robot learn about its
  environment, relying only on internally available
  information, including the sensor data? 
  Are there mathematical conditions on the internal robot
  system which can be internally verified and
  make the robot's internal system mirror the structure
  of the environment? We prove that sufficiency
  is such a mathematical principle, and mathematically describe the emergence of the robot's internal structure isomorphic or bisimulation equivalent to that of the environment.
  A connection to the free-energy principle
  is established, when sufficiency is interpreted as 
  a limit case of surprise minimization. As such, we show that
  surprise minimization leads to having an internal model isomorphic to the environment. This also parallels
  the Good Regulator Principle which states that controlling
  a system sufficiently well means having a model of it.
  Unlike the mentioned theories, ours is discrete,
  and non-probabilistic.
\end{abstract}
\section{Introduction}

From an algorithmic perspective, we are familiar with internal robot models used for motion planning, navigation, and so on, represented with coordinates in $\mathbb{R}^n$ and geometric primitives.  We have also witnessed the rapid rise of deep learning approaches to robotics, which encode internal models as parameters in a neural net, tuned from copious amounts of input-output data.  Such systems still require extensive training with vast amounts of supervisory data from human experts and numerous reinforcement learning trials in virtual environments~\cite{DRLKai}. Despite this, the derived robots remain unable to function effectively in unfamiliar environments~\cite{DRLPablo}, unlike humans who can adapt to new situations with ease. According to~\cite{LeCun2022}, this discrepancy may be
due to the ability of humans to develop internal models of how the world works. Currently, these internal models are often pre-defined for training environments, making it difficult for robots to operate in different settings. Therefore, it is crucial to understand how robots can independently construct
internal models that correlate with their external environments in a useful way without relying on prior assumptions.

To address this challenge, we explore fundamental questions: What does it mean for a robot to explore its environment? How should we conceptualize it? What does it mean for a robot to have a useful internal structure
in relationship to its external environment? These questions are fundamental to advancing robotics as they address the core challenges of enabling robots to develop a deeper understanding and adaptability to their surroundings.

\subsection{Connection to other disciplines and related work}

The relationship between the structure of an internal model and that of the external environment has been studied in the context of control theory.
In~\cite{ConAsh70} it was shown that an effective regulator of a given system needs to contain an isomorphic copy of the system itself. In~\cite{FraWon76} this statement is formulated in the context of linear, time-invariant, finite-dimensional, multi-variable control systems.
The incorporation of an internal model was formalized in terms of the divisibility of the invariant factors of the matrix describing the (linear) dynamics of the compensator by the minimal polynomial of a matrix representing the dynamics of the exogenous signals affecting the plant.
It was concluded that the synthesis of a plant and a compensator (controller) is structurally stable only if the compensator incorporates an internal model of the plant dynamics.

Meanwhile, for biophysics and cognitive systems, the field of \emph{active inference}~\cite{ParPezFri22} seeks to interpret the notions of agent, mind, and brain through the general framework of the free-energy principle~\cite{Fri12,RamFriHip20}. Its stated aim is to explain behavior of agents through the over-arching objective of surprise minimization via action. A recent work~\cite{HeiMilDaCManFriCou24} illustrates how collective behavior can emerge from the simultaneous updating of each agent's internal variables, guided by the minimization of surprise (indirectly via the minimization of free energy). However, the internal variables have pre-assigned relationships with the environment (local distance between the agent and its neighbors) governed by pre-assigned dynamics.

In the context of discrete-time stochastic dynamical systems, the problem of discovering economical, predictive representations based solely on action-observation dynamics has been formalized in the theory of \emph{predictive state representations (PSR)}~\cite{LitSut01,MaTanChePanZen21}. Based on the earlier, deterministic framework of \emph{diversity-based inference}~\cite{RivSch94}, PSR suggests a way to model such systems through maintaining a vector of probabilities for the likelihood of observing certain finite-length action-observation sequences, called core tests. Maintaining a relatively low-dimensional vector of such probabilities turns out to be sufficient for effectively modelling the statistical behaviour of the unknown dynamics.

From the perspective of classical logic, one typically deals with a language $L$ (such as first-order logic) and $L$-model $M$ with domain $\dom(M)$. A crucial aspect of the model involves functions that map elements of $\dom(M)$ to constant symbols in $L$, subsets of $\dom(M)^n$ to $n$-ary relation symbols in $L$, and functions from $\dom(M)^n$ to $\dom(M)$ to $n$-ary functions
in~$L$. This mapping is the semantic function and assigns meaning to elements of the language~$L$, relating them to elements of the world. This models, in particular, our understanding of natural language. For example, the word ``truck'' represents an actual object in the world, and as humans, we inherently understand this connection. Regardless of whether we can explain the origin or the ``reality'' of this connection, we can still formalize it using semantic functions. By formalizing these connections, we can better understand and analyze the structure of languages and theories. Consequently, these semantic functions allow us to introduce the following informal definition: \emph{A logical theory or a theory of agency is representationalist if and only if it relies on semantic functions}. In this sense, all logical theories
-- from classical first-order and second-order logic to intuitionistic logic and dependence logic
-- are representationalist. 

When the internal structure (of a human or robot ``brain'') is postulated to 
bear content or represent external states, the theory doing so is
called \emph{representational}.
For example, in robotics, the internal state is
often understood as serving as a \emph{representation}%
\footnote{That is, a correspondence between
the properties or elements of the internal model with those
of the external environment is assumed either implicitly
or explicitly.} 
of its external environment. However, the mapping, the \emph{semantic function}, which assigns meanings to this representation to relate it to elements of the external environment is often overlooked. Theoreticians or designers frequently assume that the environment is structured in a certain way and that sensor data reflects that structure in some pre-defined manner~\cite{KnownEnvKantaros,KnownEnvJana}. This assumption helps them to infuse the symbols in the robot's internal structure with meanings. Nonetheless, such an assumption limits the applicability of many robotic solutions to more realistic scenarios, such as search and rescue tasks, where the environment is previously \textit{unknown}. 
Navigating an \textit{unknown} environment presents significant challenges. A high degree of complexity and autonomy is required for the robot to make decisions based on the data it observes~\cite{autonomy2020}. To design a robot capable of autonomous learning, we should separate the meaning assigned by the designers to the labels in the robot's brain from the meaning the \emph{robot itself} assigns to them. For the robot to be truly flexible and able to generalize to unseen environments, it should be capable of learning meanings based on its own histories of interaction.

Various semantic mappings were explicitly analysed in recent work in the context of perception engineering and robotics by a subset of the authors~\cite{LaValle2023}. Understanding the nature of semantic functions is essential for developing robots capable of autonomous learning and flexible adaptation. 
In this paper, we make the following philosophical proposition: \emph{a semantic function is inherently arbitrary}. By this, we mean that the semantic function does not arise from structure in some non-arbitrary or systematic way. 
As an example of structurally non-arbitrary assignment is
the assignment of the fundamental group to a topological space.
Quoting the linguist Ferdinand de Saussure, \textit{``the bond between the signifier and the signified is arbitrary''}~\cite{SignArthur,SignSaussure}. In Saussure's theory, the ``signifier'' is the sound-image (or the form of a word), and the ``signified'' is the concept (or the meaning associated with that word). The quote describes the relationship between the signifier and the signified as an arbitrary relation, meaning that there is no inherent connection between the form of a word and its meaning; rather, this relationship is established by convention within a language community.

Enactivism, a branch of (philosophy of) cognitive science, argues that
postulating representations and semantic content is often philosophically
questionable and practically useless \cite{gallagher2017enactivist,HuttoMyin2012,huttomyin2,varela1992embodied}. 
This argument could be extended
to robotics in the light of the above discussion. Enactivism argues
that the complex structure of the brain is there not to represent
the world around us, but rather to make our engagement with it more
efficient. A counterargument is that building elaborate representations
could actually help with efficient engagement. There are two
counter-counterarguments. First is that elaborate represenations
might be useful, but they might also be unnecessary. Is it possible
to circumvent elaborate representations and go directly to efficient
engagement? The second is that when some internal configuration has
structural similarities to the environment, it does not mean it is 
a representation. In our context specifically we see (Corollary~\ref{cor:Main1})
that such internal structures emerge even when it is not the intention
or the goal of the system to generate them. For more on the second
counter-counterargument from a philosophical 
perspective, see~\cite{HuttoMyin2012,huttomyin2}.


\subsection{Contribution}

We develop a mathematical framework which enables robots to learn an internal model of the external environment by relying solely on quantities which can 
be internally evaluated. This can be seen as 
\emph{emergent structural semantics}. The robot only
engages in resolving internal ``conflicts'' such as 
minimization of surprise. 

We mathematically
prove that resolving these internal conflicts
invariably results in the emergence of an internal
structure that is
structurally similar to the environment. The main statement capturing
this intuition is Theorem~\ref{thm:Main1}.

One application 
even
shows that sometimes only proprioceptive data is enough 
for the emergence of an internal structure which is isomorphic
to the external state space, even though no other sensor data
is available, see Remark~\ref{rem:Proprio} and Example~\ref{ex:RobotArm}.
An example of ``embodied cognition''.

In this regard, our work can be viewed 
as a first step towards a discrete, combinatorial, non-probabilistic formulation of the free-energy principle (FEP) \cite{Fri05,Fri10,Fri12,ParPezFri22}. The FEP 
has gained significant attention as a model of the brain and ``living systems'' in general.
It has two principal components.
One is the analysis of the independence
of a ``living system'' from the rest
of the world, formalized through
Markov blankets. Another component
is the idea of minimization
of surprise by the system and the emergent
internal structure which results from 
it. This aspect of the FEP is the one we ``replicate'' in this paper. Unlike most
available accounts on the FEP, our framework
is not probabilistic, but rather combinatorial and in line with the previous work on history information spaces
\cite{SakWeiLav22,WeiSakLav22}.
Our results are also limit-cases
in which the surprise in not just minimized, but actually brought to zero. The result can also be seen as a combinatorial and deterministic version 
of the Good Regulator Theorem presented in
\cite{ConAsh70}.

\section{Surpriseless couplings and bisimulation}

The main tenet of the FEP that we focus on
is surprise minimization. The notion of
sufficiency 
is the quintessential notion of surprise minimization 
in the framework of information transition systems~\cite{Lav06,SakWeiLav22,WeiSakLav22}. By definition,
sufficiency is a condition posed on an equivalence relation
over a dynamical system which states that each 
equivalence
``predicts'' the next equivalence class. This allows to transfer the notion of surprise minimization from
the probabilistic context (where it is measured by entropy)
to the context of countable discrete dynamical systems and
transition systems. In this section we introduce basic definitions and the first result stating a connection between surprise minimization
and equivalence (in this case bisimulation equivalence)
of the internal state space with the external state space.

We will fix $U$ and $Y$ to be the sets of motor commands
and sensor data, respectively.

\begin{Def}
  A \emph{deterministic transition system} (DTS) is a triple
  $\SS=(S,A,\tau)$ where $\tau\colon S\times A\to S$ is the transition
  function.  A \emph{labeled} DTS is $(S,A,\tau,h)$ where $(S,A,\tau)$
  is DTS and $h$ is some labeling function.  A (labeled)
  \emph{sensorimotor transition system} is a DTS with $A=U\times Y$
  and a possible labeling function $h$ with range in~$Y$.  A 
  \emph{(labeled) internal system} is a (resp. labeled) sensorimotor transtion system with $S=I$
  the \emph{internal state space} and $\tau=\f$ the \emph{information
    transition function}. 
  The \emph{external system}, or \emph{environment} is a labeled DTS with $S=X$, $A=U$,
  $\tau=f$ and the range of $h$ being~$Y$. 
  Both external and internal systems are special cases of sensorimotor transition 
  systems. 
\end{Def}

One may ask why is the input to the internal system $U\times Y$,
and not just~$Y$. Typically we expect the behaviour of a robot
to depend on the history of sensor data, but not on its own motor commands
which are determined by the sensor data anyway. 
We do this for our framework to be general enough. 
For example,
in a situation where the robot's policy has not been determined yet,
all possible sequences of motor commands are possible. Or perhaps
the DTS is a model of only a part of the robot, a part which does 
not (yet) decide the motor commands and has to be prepared for 
all possibilities. An internal system is \emph{exploratory},
if it is prepared to ``try'' all possible motor commands indpendently
of the sensor data. Mathematically, $(I,U\times Y,\f)$ is exploratory,
if the value of $\f(\iota,u,y)$ is not dependent on~$y$, i.e.
for all $y,y'\in Y$ we have $\f(\iota,u,y)=\f(\iota,u,y')$.
For exploratory internal systems we may drop the component $Y$
from the definition. This leads to:

\begin{Def}\label{def:Exploratory}
    An \emph{exploratory internal system} is a triple
    $(I,U,\f)$ in which $I$ is the internal state space,
    and $\f\colon I\times U\to I$.
\end{Def}

Since we are interested in constructing internal systems
and not in finding policies, we will deal mostly with
exploratory internal systems. We justified the title ``exploratory''
by the intuition that the agent is ``curious'' and tries all possible
paths notwithstanding the sensor data. 
Our theorems about emergence of bisimulation equivalence 
and isomorphism will justify
this title formally because they work precisely for such systems.
Another reason to introduce the class of exploratory systems
is to simplify some of the mathematical exposition.

\begin{Def}\label{def:FreeDTS}
  Given a set $A$, the \emph{free DTS} generated by $A$, denoted
  $\FF(A)$, is the DTS $(A^{<\N},A,\cat)$ where the state space
  $A^{<\N}$ is the set of finite sequences of elements of $A$, and the
  state-transition function is the concatenation of sequences.
  When $A=U$, this DTS is the collection of all possible 
  actuator sequences. If $A=U\times Y$, then it is the 
  collection of all actuation-observation, or sensorimotor,
  sequences and is equivalent to the history information
  space $\II_{hist}$ of \cite{SakWeiLav22}, see also~\cite{Lav06}.
\end{Def}

\begin{Def}[Universal systems]\label{def:specialcases}
    The system $\FF(U\times Y)$, the free DTS generated by
    the set $U\times Y$ is the \emph{universal information transition system}
    (with respect to $U$ and~$Y$).
    Every connected internal transition system is a quotient 
    of $\FF(U\times Y)$, see Corollary~\ref{cor:Universality}.     
    The \emph{universal exploratory system} is $\FF(U)$.
    Every connected exploratory internal system is its quotient.
\end{Def}

\begin{Def}\label{def:star_op}
    For a DTS $(S,A,\tau)$, $s\in S$, and $\bar a\in A^{<\N}$,
    define $s*\bar a$ by induction on the length of $\bar a$.
    If $\bar a=\es$, then $s*\bar a=s$, and if $s*\bar a$
    is defined and $a\in A$, then $s*(\bar a\cat a)=\tau(s*\bar a,a)$.
\end{Def}

\begin{Def}\label{def:Coupling}
  Given an internal system $\II=(I,U\times Y,\f)$ and an external
  system $\XX=(X,U,f,h)$, define the coupled system
  $\XX\star\II$
  to be the DTS $(X\times I,U,g)$ where
  $g\colon X\times I\times U\to X\times I$ is defined by
  \begin{equation}\label{eq:Coupling1}
    g(x,\iota,u)=(f(x,u),\f(\iota,u,h(x))). 
  \end{equation}
  For exploratory systems, if the component $Y$ is not written,
  the equation \eqref{eq:Coupling1} becomes
  $g(x,\iota,u)=(f(x,u),\f(\iota,u)).$
\end{Def}

In view of Definition~\ref{def:Coupling}, $U$ and $Y$ constitute the
``interface'' between the internal and external.

\begin{Def}
  Given a finite sequence $\bar u=(u_1,\dots,u_n)\in U^n$
  and states $x\in X$, $\iota\in I$, there are unique
  sequences $\bar x=(x_1,\dots,x_n)$ and
  $\bar\iota=(\iota_1,\dots,\iota_n)$ such that $x_1=x$, $\iota_1=\iota$
  and for each $k<n$, $g(x_k,\iota_k,u_k)=(x_{k+1},\iota_{k+1})$.
  Denote this $\bar x$ by $x*\bar u$ and $\bar \iota$
  by $\iota\diamond\bar u$. 
  
  Note that the definition of
  $x*\bar u$ here coincides with Definition~\ref{def:star_op}.
  This is why we use the same notation for it. But the operation
  $\diamond$ on $I$ requires the coupling to be defined.  
  If the external system and/or the initial states need to be specified,
  we denote $\iota\diamond\bar u=\iota\diamond_\XX\bar u=\iota\diamond_{\XX,x,\iota}\bar u$
\end{Def}

\begin{Remark}
  If we view $U^{<\N}$ as the free monoid under concatenation, then
  $x\mapsto x*\bar u$ and $\iota\mapsto \iota\diamond\bar u$ are
  actions of this monoid on $X$ and $I$, respectively.
\end{Remark}

The coupling restricts the internal system by removing
the impossible sensorimotor sequences. More precisely:

\begin{Def}\label{def:Restr}
    Suppose $\II=(I,U\times Y,\f)$ is an internal system
    and $X=(X,U,f,h)$ external, and let $(x_0,\iota_0)\in X\times I$.
    Let $\II\rest_{\XX,x_0,\iota_0}$ be the 
    DTS $(I,U,\psi)$ where $\psi(\iota,u)=\iota\diamond_{\XX,x_0,\iota_0} u$.
    If the initial states are clear from the context, we drop them from
    the subscript.
\end{Def}

\begin{Def}[Surpriseless]
  Suppose $\II=(I,U\times Y,\f)$ and $\XX=(X,U,f,h)$ are internal and
  external systems, respectively. We say that $(x_0,\iota_0)\in X\times I$
  is \emph{surpriseless}, if there are no sequences
  $\bar u=(u_1,\dots,u_{n})\in U^n$ and
  $\bar u'=(u'_1,\dots,u'_{m})\in U^m$ such that
  $\iota_0\diamond\bar u=\iota_0\diamond \bar u'$, but
  $h(x_0*\bar u)\ne h(x_0*\bar u')$.
\end{Def}

A surprise occurs if the sensory data cannot be ``predicted'' based on
the robot's internal state. Note that the condition of
being surpriseless can be evaluated within the internal model because
the definition speaks only about internal states and sensory data.

\begin{Thm}
    The coupling of $\FF(U\times Y)$
    to any environment 
    $\XX=(X,U,f,h)$ is surpriseless.
\end{Thm}
\begin{proofV}
   In the universal automaton,
   any two sequences $\bar u,bar u'$ yield a different internal state
   $\iota_0\diamond \bar u\ne\iota_0\diamond$
   if and only if $\bar u\ne \bar u'$.
   But if $\bar u=\bar u'$,
   then also $x_0*\bar u=x_0*\bar u'$,
   so we have checked the definition.
\end{proofV}

\begin{Def}[Bisimulation]
  Suppose $\II=(I,U\times Y,\f,h')$ is a labeled internal system.  
  Suppose $\XX=(X,U,f,h)$ is an external system. We
  say that $x_0\in X$ is \emph{bisimulation equivalent to
  $\iota_0\in I$ in $\XX\star\II$}, if there is a relation
  $R\subseteq X\times I$ such that
  (B1.) $(x_0,\iota_0)\in R$,
  (B2.) for all $(x,\iota)\in R$ and all $u\in U$, $(x* u,\iota \diamond u)\in R$, and (B3.) for all $(x,\iota)\in R$ we have $h(x)=h'(\iota)$.
\end{Def}

At first glance, bisimulation equivalence cannot be evaluated
within the internal system because by definition it requires
finding a binary relation between the internal and the external.
In Theorem~\ref{thm:SurpriselessIsBisim} below we prove, however,
that it is equivalent to being surpriseless.

\begin{Def}\label{def:StronglyConnected}
  A DTS $(S,A,\tau)$ is \emph{strongly connected}, if for all
  $s,s'\in S$ there is $\bar a\in A$ with $s*\bar a= s'$.
\end{Def}

\begin{Lemma}\label{lem:Surpriseless}
  Suppose the coupled system $\XX\star \II$ is strongly connected.
  If $(x_0,\iota_0)\in X\times I$ is surpriseless, then there is a
  well-defined $\hat h\colon I\to Y$ such that
  $\hat h(\iota)=h(x_0*\bar u)$ for some (all) $\bar u\in U^{<\N}$
  with $\iota_0*\bar u=\iota$.
\end{Lemma}
\begin{proofV}
  It is enough to show that for all $\bar u,\bar u'\in U^{<\N}$, if
  $\iota_0\diamond\bar u=\iota_0\diamond\bar u'$, then
  $h(x*\bar u)=h(x*\bar u')$, but this is exactly the
  definition of surpriseless.
\end{proofV}

\begin{Thm}\label{thm:SurpriselessIsBisim}
  Suppose the coupled system $\XX\star \II$ is strongly connected.
  The following are equivalent
  \begin{enumerate}
  \item $(x_0,\iota_0)$ is surpriseless in $\XX\star\II$
  \item $\hat h$ is well-defined
  \item $\hat h$ is well-defined and $x_0$ is bisimulation equivalent
    to~$\iota_0$ in $\XX\star\II$ over~$\hat h$.
\end{enumerate}
\end{Thm}
\begin{proofV}
  We already proved $1\to 2$ (Lemma~\ref{lem:Surpriseless}).  Assume 2. Let
  $$R=\{(x,\iota)\in X\times I\mid \exists\bar u\in U^{<\N}((x,\iota)=(x_0*\bar u,\iota\diamond\bar u))\}.$$
  Conditions (B1) and (B2) are clearly satisfied.  If $(x,\iota)\in R$, then
  $h(x)=h(x_0*\bar u)=\hat h(\iota)$ by definition of $\hat h$
  which proves also condition~(B3). This proves $2\to 3$. Clearly $3\to 2$.
  Suppose that 2 and we will prove 1. 
  Suppose $\bar u,\bar u'$ are such that $\iota_0\diamond\bar u=\iota_0\diamond\bar u'$.
  Since $\hat h$ is well-defined, we have
  $h(x_0*\bar u)=\hat h(\iota_0\diamond\bar u)=\hat h(\iota_0\diamond\bar u')= h(x_0*\bar u')$. 
\end{proofV}

Theorem~\ref{thm:SurpriselessIsBisim} shows that if the agent had a 
method to reduce surprise,
then once successful, its internal system would be 
bisimulation equivalent to the environment.
This result is reminiscent of the Good Regulator \cite{ConAsh70}. Surpriselessness
implies the theoretical possibility to control the system completely. In 
a surpriseless system, the internal state determines the results of all possible
future actions (see also Lemma~\ref{lemma:Sufhattau}). On the other hand we just
proved that such potential to control the system implies bisimulation equivalence.
Bisimulation equivalence, however, is significantly weaker than isomorphism.
Also, Theorem~\ref{thm:SurpriselessIsBisim} is non-constructive,
it does not say whether or when surpriseless couplings exist, or how
to obtain them. These questions will be addressed in the next section.

\section{Theory for General Deterministic Transition Systems}

In the previous section we showed that if an internal state space is adapted to the
environment with the goal of minimizing surprise, then it will become bisimulation equivalent to it. What are the conditions under which surprise minimization
leads to the internal system being isomorphic, and
not just bisimilar, to the external system?
A bisimulation equivalence between transition systems is an
isomorphism, if both systems are 
minimal in some sense.
The robot cannot know whether the environment is
in this sense minimal, but it can ``know'' when its internal system is. 
It is cumbersome to speak about the function $\hat h$ of Lemma~\ref{lem:Surpriseless} 
because it is
not always well-defined. Also, Theorem~\ref{thm:SurpriselessIsBisim}
does not shed light on how to systematically obtain surpriseless
internal systems when nothing is known about the environment or when
do they even exist. We will
now present a framework which enables a more systematic
characterization of when the internal system is both surpriseless and
isomorphic to the external system, and how to obtain them, theoretically,
but not algorithmically.%
\footnote{Our result shows how to find such systems by computing minimal sufficient
refinements on infinite trees. As such this is not computationally feasible,
but gives a theoretical doorway towards designing algorithms.}
This approach will pave the path to algorithm
design in future work, see Section~\ref{sec:Discussion}.

\begin{Def}\label{def:Suff}
  Given a DTS $\SS=(S,A,\tau)$.  We say that an equivalence relation
  $E$ on $S$ is \emph{sufficient}, or $\SS$-sufficient, if for all
  $s,s'\in S$ and $a\in A$ we have that $(s,s')\in E$ implies
  $(\tau(s,a),\tau(x',a))\in E$. An equivalence relation $E'$ is a
  \emph{refinement} of an equivalence relation $E$, if
  $E'\subseteq E$. We say that an equivalence relation $E'$ is a
  \emph{minimal sufficient refinement} of $E$, if $E'$ is sufficient,
  $E'\subseteq E$ and for all sufficient refinements $E''\subseteq E$ we
  have $E''\subseteq E'$.
\end{Def}

The following is Theorem 4.19 of \cite{WeiSakLav22}.

\begin{Thm}\label{thm:MSRUnique}
  Suppose $\SS=(S,A,\tau)$ is a DTS. Let $E$ be an equivalence relation
  on $S$. Then the minimal $\SS$-sufficient refinement of $E$ 
  exists and is unique. We denote it
  by~$\MSR(E)=\MSR_{\SS}(E)$. \qed
\end{Thm}

\begin{Def}
  Let $\SS_i=(S_i,A,\tau_i)$ be a DTS for $i\in\{0,1\}$.  A
  function $h\colon S_0\to S_1$ is called a \emph{homomorphism from
    $\SS_0$ to $\SS_1$}, if for all $(s,a)\in S_0\times A$ we have
  $$\tau_1(h(s),a)=h(\tau_0(s,a)).$$
  It is called an \emph{epimorphism}, if it is onto, and
  \emph{isomorphism} if it is a bijection.
\end{Def}

\begin{Def}\label{def:Invariant}
  Let $E$ be an equivalence relation on $S$. If $s\in S$, denote by
  $[s]$, or by $[s]_E$, the equivalence class of $s$ which is the set
  $\{s'\in S\mid (s,s')\in E\}$. The set of all equivalence classes is
  denoted by~$S/E$. 
  Suppose that $h\colon S\to Y$ is a function and
  $E$ is an equivalence relation on~$Y$. Define
  $$h^{-1}(E)=\{(s,s')\in S\times S\mid (h(s),h(s'))\in E\}.$$
  It is easy to verify that $h^{-1}(E)$ is an equivalence relation
  on~$S$.
  Denote by $E_h$ the equivalence relation
  $h^{-1}(\id_{Y})$. Equivalently, $(s,s')\in E_h\iff h(s)=h(s')$.
  A function $h\colon S\to Y$ is $E$-\emph{invariant},
  if $E$ is a refinement of~$E_h$.
  We say that an equivalence relation $E$ is \emph{$h$-closed},
  if $E_h$ is a refinement of~$E$.
\end{Def}

We leave the following two observations for the reader to prove:

\begin{Lemma}\label{lemma:Ehclosedinv}
    Let $\SS=(S,A,\tau)$ be DTS and $h\colon S\to Y$.
    Then $E_h$ is $h$-closed, and $h$ is $E_h$-invariant. \qed
\end{Lemma}

\begin{Lemma}\label{lem:SuffWD}
  Suppose $\SS=(S,A,\tau)$ is a DTS, $E$ is an
  $\SS$-sufficient equivalence relation, and $h\colon S\to Y$.
  If $E$ is $h$-closed, then $(h/E)\colon \SS/E\to Y$ 
  defined by $(h/E)([s])=h(s)$ is well-defined. \qed
\end{Lemma}

\begin{Lemma}(Lemma 4.5 of \cite{WeiSakLav22})
   \label{lemma:Quotient}
  Suppose $\SS=(S,A,\tau)$ is a DTS, and $E$ is an
  $\SS$-sufficient equivalence relation.  Let $S'=S/E$ and define
  $\tau'\colon S'\times A\to S'$ by $\tau'([s],a)= [\tau(s,a)]$
  (also denoted $\tau'=\tau/E$).  Then $\tau'$ is well-defined and
  $\SS'=(S',A,\tau')$ is a DTS (also denoted by $\SS'=\SS/E$). \qed
\end{Lemma}

\begin{Prop}[Pullback]\label{prop:EpimorphPull}
  Suppose $\SS_i=(S_i,A,\tau_i)$ is a DTS for $i\in \{0,1\}$.
  Suppose that $E_1$ is an $\SS_1$-sufficient equivalence relation on
  $S_1$.  Suppose that $h\colon S_0\to S_1$ is an epimorphism.  
  Then $E_0=h^{-1}(E_1)$ is an $\SS_0$-sufficient equivalence relation on $S_0$ and
  $\SS_0/E_0\cong\SS_1/E_1$.
\end{Prop}
\begin{proofV}
  Suppose $s,s'\in S_0$ are $E_0$-equivalent and suppose $a\in A$.
  Then we have $h(s)\,E_1\, h(s')$ and so by the sufficiency of $E_1$
  we have $\tau_1(h(s),a)\,E_1\,\tau_1(h(s'),a)$. On the other hand because
  $h$ is a homomorphism, $h(\tau_0(s,a))=\tau_1(h(s),a)$ and
  $h(\tau_0(s',a))=\tau_1(h(s'),a)$. Combining these, we have that
  $h(\tau_0(s,a))$ and $h(\tau(s',a))$ are $E_1$-equivalent. By the definition of $E_0$,
  this means that $\tau_0(s,a)$ and $\tau_0(s',a)$ are $E_0$-equivalent which proves that $E_0$ is $\SS_0$-sufficient.

  Now we prove that $\SS_0/E_0\cong \SS_1/E_1$. To simplify
  notation, for all $i\in \{0,1\}$, denote $[s]_i$ instead of
  $[s]_{E_i}$ , and also $S_i'$ and $\tau_i'$ instead of
  $S_i/E_i$ and $\tau_i/E_i$.
  Define $\hat h\colon S'_0\to S'_1$ by
  \begin{equation}
    \hat h([s]_0)=[h(s)]_1\label{eq:WD}    
  \end{equation}
  By the definition of $E_0$ it is easy to see that $\hat h$ is
  well-defined. Suppose that $s\in S_0$ and $a\in A$ are arbitrary.
  Then
  \begin{align*}
    &\hat h(\tau_0'([s]_0,a))
    \stackrel{(a)}{=}\hat h([\tau_0(s,a)]_0)
    \stackrel{(b)}{=}[h(\tau_0(s,a))]_1 
    \stackrel{(c)}{=}[\tau_1(h(s),a)]_1 
    \stackrel{(d)}{=}\tau_1'([h(s)]_1,a)\\
    \stackrel{(e)}{=}&\tau_1'(\hat h([s]_0),a) 
  \end{align*}
  Here (a) and (d) follow from the definitions of $\tau_0'$ and $\tau_1'$,
  (b) and (e) from \eqref{eq:WD}, and (c) from that $h$ is a homomorphism.
  
  This shows that $\hat h$ is a homomorphism from $\SS_0/E_0$ to
  $\SS_1/E_1$. It remains to show that $\hat h$ is a bijection.
  Since $h$ is onto (by the definition of an epimorphism), it is
  standard to check that $\hat h$ is also onto.  Suppose $s,s'\in S_0$
  are not $E_0$-equivalent.  Then by the definition of $E_0$ we have
  that $h(s)$ and $h(s')$ are not $E_1$-equivalent, so by
  \eqref{eq:WD} we have $\hat h([s]_0)\ne \hat h([s']_0)$ which proves
  that $\hat h$ is one-to-one. Thus we have proved that $\hat h$ is a
  bijective homomorphism, i.e. an isomorphism, from $\SS_0/E_0$
  to $\SS_1/E_1$, so the proof is complete.
\end{proofV}

\begin{Cor}[Universality]\label{cor:Universality}
  Let $\FF(A)$ be as in Definition~\ref{def:FreeDTS}.
  The free DTS of is \emph{universal} in the sense that if $\SS=(S,A,\tau)$
  is any strongly connected DTS, then there is an equivalence relation
  $E$ on $A^{<\N}$ such that $\FF(A)/E\cong \SS$.
\end{Cor}
\begin{proofV}
  Fix $s_0\in S$.
  Define the map $h\colon A^{<\N}\to S$ by induction on the length of
  $\bar a$ as follows. Let $h(\es)=s_0$, and if $h(\bar a)$ is defined,
  let $h(\bar a\cat a)=\tau(h(\bar a),a)$. 
  Let $E_1=\id_S$. Then it is clearly $\SS$-sufficient.
  Let $E_0=h^{-1}(E_1)$. By strong connectedness, $h$ is an epimorphism.
  Now the result follows
  from Proposition~\ref{prop:EpimorphPull}.
\end{proofV}

Corollary~\ref{cor:Universality} means that finding a DTS is
equivalent to finding an equivalence relation on $\FF(A)$.  It follows
that finding an appropriate internal system for a robot's brain can be
reformulated as finding such an equivalence relation. The condition of
sufficiency is essentially the same as surprise minimization: by
definition of sufficiency the equivalence class of a state predicts
the equivalence class of the following state. This condition can be
evaluated within the internal system, but we will show analogously to
Theorem~\ref{thm:SurpriselessIsBisim} that if the robot can achieve
sufficiency, then it achieves equivalence. In fact, minimal
sufficiency will imply isomorphism between the internal and external
systems under some conditions.

\begin{Def}
  Suppose $H\subseteq S\times S$ is any set. Then
  $\la H\ra$ is the \emph{equivalence relation generated by~$H$}. It is defined
  to be the set of all pairs $(s_0,s_1)\in S\times S$ such that
  there exists a finite sequence $(x_0,\dots,x_k)$ of elements of $S$
  such that $s_0=x_0$, $s_1=x_k$, and for all $i\in \{0,\dots,k-1\}$ we have $(x_i,x_{i+1})\in H$ or $(x_{i+1},x_i)\in H$.
\end{Def}

\begin{Lemma}\label{lemma:Union}
  Suppose $\SS_i=(S_i,A,\tau_i)$ is DTS for $i\in \{0,1\}$,
  $E,E^0,E^1,\dots,E^{m-1}$ are equivalence relations on $S_0$, $E_1$
  an equivalence relation on $S_1$, and $h\colon S_0\to S_1$ a
  function. Then the following hold:
  \begin{enumerate}
  \item Each $E^i$ is a refinement of $\la E^0\cup \cdots\cup E^{m-1}\ra$, \label{longlemma:1}
  \item If $E^i$ are $\SS_0$-sufficient for all $i<m$, then so is
    $\la E^0\cup\cdots\cup E^{m-1}\ra$,
    \label{longlemma:2}
  \item If $E^i$ is a refinement of $E$ for all $i<m$, then
    $\la E^0\cup \cdots\cup E^{m-1}\ra$ is also a refinement of~$E$.
    \label{longlemma:3}
  \item If $E$ is a refinement of $E^i$ for all $i<m$, then
    $E$ is a refinement of
    $\la E^0\cup \cdots\cup E^{m-1}\ra$.
    \label{longlemma:4}
  \item If $E^i$ is $h$-closed for all $i<m$,
    then so is $\la E^0\cup \cdots\cup E^{m-1}\ra$.
    \label{longlemma:5}
  \item $E_h$ is a refinement of $h^{-1}(E_1)$.
    \label{longlemma:6}
  \item If $h$ is onto, and $E$ is $h$-closed, then
    $$h(E)=\{(h(s),h(s'))\in S_1\times S_1\mid (s,s')\in E\}$$
    is an equivalence relation on $S_1$.
    \label{longlemma:7}
  \item If $h$ is onto, $E$ is $h$-closed, and $E$ is a refinement of $h^{-1}(E_1)$,
    then $h(E)$ is a refinement of~$E_1$.
    \label{longlemma:8}
  \item If $h$ is an epimorphism, $E$ is $\SS_0$-sufficient and  $h$-closed, 
    then $h(E)$ is $\SS_1$-sufficient.
    \label{longlemma:9}
  \item If $h$ is a homomorphism and $E_1$ is $\SS_1$-sufficient, then $h^{-1}(E_1)$
    is $\SS_0$-sufficient,
    \label{longlemma:10}
  \item If $h$ is a homomorphism, then $E_h$ is $\SS_0$-sufficient.
    \label{longlemma:11}
  \end{enumerate}
\end{Lemma}
\begin{proofV}
      From definitions it follows that
  $$E^i\subseteq E^0\cup\cdots\cup E^{m-1}\subseteq  \la E^0\cup\cdots\cup E^{m-1}\ra$$
  and by definition of a refinement we have~\ref{longlemma:1}.
  \ref{longlemma:2} is Theorem 4.15 of \cite{WeiSakLav22}. Suppose
  $(s_0,s_1)\in \la E^0\cup \cdots\cup E^{m-1}\ra$ and
  $x_0,\dots,x_{k-1}$ witness this. Then by the assumption, each pair
  $(x_{i},x_{i+1})$ is in $E$. By transitivity of $E$ then also
  $(s_0,s_1)\in E$.  This proves \ref{longlemma:3}.  \ref{longlemma:4}
  follows easily from~\ref{longlemma:1}.  By definition of being
  $h$-closed, \ref{longlemma:5} follows directly from
  \ref{longlemma:4}. Suppose $(s,s')\in E_h$. Then $h(s)=h(s')$, so by
  reflexivity of $E_1$, $(h(s),h(s'))\in E_1$ which proves  \ref{longlemma:6}.
  For \ref{longlemma:7},
  reflexivity follows from the surjectivity of~$h$.  Symmetry follows
  from the symmetry of $E$. For transitivity, assume
  $(s_1,s_2)\in h(E)$ and $(s_2,s_3)\in h(E)$. Let
  $s_1',s_2',s_2'',s_3'$ be some elements of $S_0$ such that
  $h(s_1')=s_1$, $h(s'_2)=h(s''_2)=s_2$, and $h(s'_3)=s''_3$, and so
  that $(s'_1,s'_2)\in E$ and $(s''_2,s'_3)\in E$. But
  $(s'_2,s''_2)\in E_h$ which is a refinement of $E$, so also
  $(s'_2,s''_2)\in E$. Now by transitivity of $E$, we have
  $(s'_1,s'_3)\in E$ and so $(s_1,s_3)=(h(s'_1),h(s'_3))\in h(E)$ and
  we are done. By \ref{longlemma:7} $h(E)$ is an equivalence 
  relation and since
  $E\subseteq h^{-1}(E_1)$ implies $h(E)\subseteq E_1$, this proves~\ref{longlemma:8}.

  For \ref{longlemma:9}, let $(s_1,s_2)\in h(E)$ and $a\in A$.  Let
  $s'_1,s'_2\in S_0$ witness that $(s_1,s_2)\in h(E)$.  Then
  $\tau(s_1,a)=\tau(h(s'_1),a)$ and $\tau(s_2,a)=\tau(h(s'_2),a)$ and
  $(s'_1,s'_2)\in E$. By the assumed sufficiency of $E$, and the
  properties of a homomorphism, we have
  $\tau(s_1,a)=\tau(h(s'_1),a)=h(\tau(s'_1,a))=h(\tau(s'_2,a))=\tau(h(s'_2),a)=\tau(s_2,a)$
  which proves the required sufficiency of~$h(E)$.  The proof of
  \ref{longlemma:10} is analogous.  \ref{longlemma:11} follows from
  \ref{longlemma:10} and the sufficiency of~$\id_{S_1}$.
\end{proofV}

The following theorem is the key to understanding why internal
processing is ``enough''. It shows that the operator $\MSR$
commutes with any epimorphism. One can compute the minimal sufficient
refinement of a relation and then take the inverse image of the
result. Or,
one can first take the inverse image, and then compute the minimal sufficient refinement
of that. According to Theorem~\ref{thm:Commute} 
both give the same result. This is valuable for us, 
because the inverse image of the relation (before applying~$\MSR$)
corresponds to what the robot can sense. Then, the robot
can internally apply~$\MSR$, and can be sure to get the same
as if $\MSR$ was applied in the external system first.
We explore the applications of Theorem~\ref{thm:Commute} to the robot
system in Section~\ref{sec:Applications}.

\begin{Thm}\label{thm:Commute}
  Let $\SS_i=(S_i,A,\tau_i)$ be DTS for $i\in \{0,1\}$, and
  suppose that $h\colon S_0\to S_1$ is an epimorphism. Suppose
  $E_1$ is an equivalence relation on $S_1$. Then
  $$\MSR_{\SS_0}(h^{-1}(E_1))=h^{-1}(\MSR_{\SS_1}(E_1)).$$
\end{Thm}
\begin{proofV}
  Denote $E_0 = h^{-1}(\MSR_{\SS_1}(E_1))$.  By
  Proposition~\ref{prop:EpimorphPull} the relation $E_0$ is
  sufficient. It is also a refinement of $h^{-1}(E_1)$ because
  $\MSR_{\SS_1}(E_1)\subseteq E_1$ implies
  $h^{-1}(\MSR_{\SS_1}(E_1))\subseteq h^{-1}(E_1)$.  By
  Theorem~\ref{thm:MSRUnique} it is now enough to show that $E_0$
  is a minimal sufficient refinement of $h^{-1}(E_1)$.  Suppose
  for a contradiction that there exists an $\SS_0$-sufficient
  refinement $E'_0$ of $h^{-1}(E_1)$ such that
  $E'_0\not\subseteq E_0$. Let
  $E'_1=h(\la E_0\cup E'_0\cup E_h\ra)$.
  By Lemma~\ref{lemma:Union}\eqref{longlemma:2} and \ref{lemma:Union}\eqref{longlemma:11} 
  the relation $\la E_0\cup E'_0\cup E_h\ra$ is $\SS_0$-sufficient, and 
  by Lemma \ref{lemma:Union}\eqref{longlemma:1} and~\ref{lemma:Union}\eqref{longlemma:4} 
  it is $h$-invariant.
  So by \ref{lemma:Union}\eqref{longlemma:9} $E'_1$ is $\SS_1$-sufficient.
  Since $E'_0\not\subseteq E_0$, 
  we have by \ref{lemma:Union}\eqref{longlemma:1} that
  $$\la E_0\cup E'_0\cup E_h\ra\not\subseteq E_0$$
  and so
  \begin{equation}
    E'_1 = h(\la E_0\cup E'_0\cup E_h\ra)\not\subseteq h(E_0)=
    \MSR_{\SS_1}(E_1).\label{eq:NotRefinement}
  \end{equation}
  On the other hand $E_0$ and $E_0'$ are refinements of $h^{-1}(E_1)$,
  so by~\ref{lemma:Union}\eqref{longlemma:3}
  and~\ref{lemma:Union}\eqref{longlemma:6}
  $\la E_0\cup E_0'\cup E_h\ra$ is a refinement of $h^{-1}(E_1)$.  By
  \ref{lemma:Union}\eqref{longlemma:8} we now have that $E'_1$ is a
  refinement of~$E_1$. Combining this with \eqref{eq:NotRefinement}
  above, and the sufficiency of $E_1'$, we have a contradiction with
  the minimality of $\MSR_{\SS_1}(E_1)$.
\end{proofV}

\begin{Lemma}\label{lemma:Sufhattau}
  If $E$ is a sufficient equivalence relation on $\SS=(S,A,\tau)$,
  then for all $\bar a\in A^{<\N}$ and all $s,s'\in S$ we have
  $$(s,s') \in E\quad \Longrightarrow\quad (s_0*\bar a, s'*\bar a)\in E.$$
\end{Lemma}
\begin{proofV}
    By induction on the length of $\bar a$.
\end{proofV}


\section{Application to Internal and External Systems}
\label{sec:Applications}

Recall that we conventionally fix $U$ and $Y$ to be the sets of motor commands
and sensor data, respectively.





In this section we will work with 
the universal exploratory 
system (Definition~\ref{def:Exploratory}) 
$\II=\FF(U)$ in which
the initial state is always the empty sequence.
Given an external system $\XX=(X,U,f,h)$, and 
initial state $x_0\in X$, 
let $\hat f_{x_0}\colon U^{<\N}\to X$ denote the function 
$\hat f_{x_0}(\bar u)=x_0*\bar u$. 
Note that $\hat f_{x_0}$ is a homomorphism from 
$\FF(U)$ to~$\XX$.

\begin{Thm}\label{thm:Main1}
  Let $\XX=(X,U,f,h)$ be a strongly connected external system with the
  initial state $x_0\in X$.
  Suppose that $E$ is any equivalence relation on~$X$.
  Let $E_{\II}=\MSR_{\FF(U)}(\hat f^{-1}_{x_0}(E))$, and
  $E_\XX=\MSR(E)$. Then
  $$\II/E_{\II}\cong \XX/E_{\XX}.$$
\end{Thm}
\begin{proofV}
  Since $\XX$ is connected, $\hat f_{x_0}$ is an epimorphism. By Theorem~\ref{thm:Commute} we have
  $$E_{\II)}=\MSR_{\FF(U)}(\hat f_{x_0}^{-1}(E))
  =\hat f^{-1}_{x_0}(\MSR_{\XX}(E))=\hat f^{-1}_{x_0}(E_\XX).$$
  The result now follows from Proposition~\ref{prop:EpimorphPull}.
\end{proofV}

It is useful to note that $\MSR(E_h)$ can be a measure of symmetry
of the environment $(X,U\times Y,f,h)$. A bisimulation
is \emph{trivial}, if it is the identity relation.
An \emph{autobisimulation} is a bisimulation
of a DTS with itself, a relation on
$X\times X$. The following is Theorem 4.4 of \cite{WeiSakLav22}.

\begin{Thm}\label{thm:Symmetry}
  An environment $\XX=(X,U,f,h)$,
  has a non-trivial autobisimulation if and only if
  $\MSR(E_h)\ne\id_{X}$. \qed
\end{Thm}

The notion of autobisimulation is a weak version of automorphism.
There is also a one-sided version of this theorem for automorphisms:

\begin{Thm}
    If there is a non-trivial automorphism of $\XX$, 
    then \mbox{$\MSR(E_h)\ne\id_X$.}
\end{Thm}
\begin{proofV}
  An automorphism is a special case of an autobisimulation,
  so the result follows from Theorem~\ref{thm:Symmetry}.
\end{proofV}

So we can call the environment \emph{chiral}, if 
$\MSR(E_h)=\id_X$.
The equivalence relation 
$\hat f^{-1}_{x_0}(E_h)$ on $\FF(U)$
equates those paths which lead to identical sensor 
data. So we may
call it the relation of \emph{sensory indistinguishability}.  The
following corollary shows that if the environment is chiral, then
taking the quotient of the history information space by the minimal
sufficient refinement of the sensory indistinguishability yields a
structure isomorphic to the environment.

\begin{Cor}\label{cor:ReversibleID}
  Suppose $\XX=(X,U,f,h)$ is a strongly connected environment,
  and $x_0\in X$.
  Suppose that
  $\MSR_\XX(E_h)=\id_{X}$. Then
  $$\FF(U)/\MSR(\hat f_{x_0}^{-1}(E_h))\cong \XX.$$  
\end{Cor}
\begin{proofV}
  Since $\XX/\id_{\XX}\cong \XX$, the result follows from
  Theorem~\ref{thm:Main1}.
\end{proofV}

Taking the minimal sufficient refinement can be interpreted as
minimizing surprise while also minimizing computational resources.
The minimization of surprise is due to the definition of sufficiency
which say that the current equivalence class predicts the next one.
Minimization of resources is captured by the minimality of the refinement,
as it yields the smallest possible quotient space.
Isomorphism is a special case of bisimulation, so 
applying Theorem~\ref{thm:SurpriselessIsBisim}
we get:
\begin{Cor}
    Suppose $\XX$ is as in Corollary
    \ref{cor:ReversibleID},
    and 
    $\II=\FF(U)/\MSR(\hat f_{x_0}^{-1}(E_h))$.
    Then $\II\star\XX$
    is surpriseless. \qed
\end{Cor}


An interesting class of chiral environments consists of
those environments in which $E_h$ has
one equivalence class which is a singleton. Call such an 
equivalence relation \emph{pointed}.
We will show that environments with pointed $E_h$
are chiral under some very minimal assumptions on~$\XX$. 
An
automaton $(S,A,\tau)$ is \emph{minimally distinguishing}, if for all
$s_0,s_1,s_2\in S$ and all $a\in A$ we have that if
$\tau(s_1,a)=\tau(s_2,a)=s_0$, then one of the following holds:
$s_0=s_1$, $s_1=s_2$ or $s_0=s_2$.

\begin{Thm}\label{thm:Pointed}
  Let $\SS=(S,A,\tau)$ be a strongly connected and minimally
  distinguishing. If $E$ is a pointed equivalence relation on
  $X$, then $\MSR(S)=\id_X$.
\end{Thm}
\begin{proofV}
   Let $s_0\in S$ be an element such that $\{s_0\}$ is an
  $E$-equivalence class. Let $s_1,s_2\in S$.
  We will show that
  $s_1\ne s_2$ implies $(s_1,s_2)\notin \MSR(E)$.
  Let $\bar a_i$ be of minimal length such that
  $s_i*\bar a_i=s_0$ for $i\in \{1,2\}$. These exist by 
  connectedness. Denote $k_i=|\bar a_i|$ for
  $i\in \{1,2\}$. Without loss of generality assume that $k_1\le
  k_2$. There are two cases: $k_1<k_2$ and $k_1=k_2$.

  \emph{Case 1:} $k_1<k_2$. By the minimality of $|\bar a_2|$, we
  have $s_2*\bar a_1\ne s_0$, but $s_1*\bar a_1=s_0$.
  By the choice of $s_0$ we have now
  $(s_2*\bar a_1,s_1*\bar a_1)\notin E,$
  and so
  $(s_2*\bar a_1, s_1*a_1)\notin \MSR(E).$
  By Lemma~\ref{lemma:Sufhattau} we now conclude that
  $(s_1,s_2)\notin \MSR(E)$.

  \emph{Case 2:} $k_1=k_2$. By induction on $k=k_1=k_2$.  We will show
  again that
  \begin{equation}
    s_1\ne s_2\ \Longrightarrow \ s_1*\bar a_1 \ne s_2* a_1\label{eq:toprove2}  
  \end{equation}
  which by the same argument as above implies that
  $(s_1,s_2)\notin \MSR(E)$.  We will proceed by induction on~$k$.

  If $k=0$, then $\bar a_1=\bar a_2$ are both empty, so $s_1=s_2=s_0$,
  and the premise $s_1\ne s_2$ is false. Suppose we have proved
  \eqref{eq:toprove2} for all $k\le n$ and that $k=n+1$. Assume
  $s_1\ne s_2$. Let $\bar a_1=a^0_1\cdots a^{n}_1$. Let $j$ be the
  smallest one such that
  \begin{equation}
    s_1*a^0_1\cdots a^j_1=s_2*a^0_2\cdots a^j_2.\label{eq:theyarethesame}
  \end{equation}
  If such $j$ does not exist, then we are done. Otherwise we will
  deduce a contradiction.  Define
  \begin{align*}
    s_1'&= s_1 * a^0_1\cdots a^{j-1}_1,\\
    s_2'&= s_2 * a^0_1\cdots a^{j-1}_1,\\
    s_0'&= s_1 * a^0_1\cdots a^{j-1}_1a^{j}_1= s_2 * a^0_1\cdots a^{j-1}_1a^{j}_1=\tau(s_1',a^{j}_1)=\tau(s_2',a^{j}_1).
  \end{align*}
  Now by the assumption that $\XX$ is minimally distinguishing we must
  now have one of the following: $(a)$ $s_0'=s_1'$, 
  $(b)$ $s_1'=s_2'$,
  or 
  $(c)$ $s_0'=s_2'$. By the minimality of $j$ we cannot have
  $(b)$. Let
  $$\bar a_1^{<j}=a^0_1\cdots a^{j-1}_1\text{ and }\bar a_1^{>j}=a^{j+1}_1\cdots a^{n}_1.$$
  If $(a)$ holds, then we have
  \begin{align*}
    s_0&= s_1 * (\bar a^{<j}_1 \cat a^j\cat \bar a^{>j}_1)
       = ((s_1 * (\bar a^{<j}_1\cat a^j))* \bar a^{>j}_1\\
       &= s_0' * \bar a^{>j}_1& s_1 * (\bar a^{<j}_1\cat a^j_1)=s_0\\
       &= s_1' * \bar a^{>j}_1& \text{by }(a)\\
       &= (s_1 * \bar a^{<j}_1) * \bar a^{>j}_1
       = s_1 * (\bar a^{<j}_1 \cat \bar a^{>j}_1)
  \end{align*}
  contradicting the minimality of~$|\bar a_1|$. If $(c)$ holds we get
  similarly a contradiction with the minimality of $|\bar a_2|$:
  \begin{align*}
    s_0&= s_0' * \bar a^{>j}_1
       = s_2' * \bar a^{>j}_1& \text{by }(c)\\
       &= (s_2 * \bar a^{<j}_1) * \bar a^{>j}_1
       = s_2 * (\bar a^{<j}_1 \cat \bar a^{>j}_1).
  \end{align*}
  This completes the proof of the theorem.
\end{proofV}

We say that a sensor mapping $h\colon X\to Y$ is \emph{pointed}, if
$E_h$ is a pointed equivalence relation which is equivalent to saying
that there is some $y\in Y$ such that $h^{-1}(y)$ is a singleton.

\begin{Cor}\label{cor:Main1}
  Let $(X,U,f,h)$ be a strongly connected minimally distinguishing
  external system. Let $x_0\in X$, and suppose that $h\colon X\to Y$
  is a pointed sensor mapping. Let
  $E=\MSR_{\FF(U)}(\hat f^{-1}_{x_0}(E_h))$.  Then
  $$\FF(U)/E\cong \XX.$$
\end{Cor}
\begin{proofV}
  By Theorem \ref{thm:Pointed}, $\MSR_{\XX}(S)=\id_\XX$. Then apply Corollary~\ref{cor:ReversibleID}.
\end{proofV}

\begin{Remark}\label{rem:Proprio}
 Suppose the state space consists of positions of the robot's
 body in the environment.
 Then it follows that the robot does not  need any ``external'' sensors
 to successfully mirror the environment internally. It is enough 
 to have proprioceptive feedback to single one specific position 
 of its own body. Striving for sufficiency (surprise minimization) takes care of the rest by Corollary~\ref{cor:Main1}.
\end{Remark}

Finally, we give an application for a mathematically
ideal situation where the motor commands generate
a group acting on the external state space:

\begin{Cor}\label{cor:GroupAction}
  Suppose that $(G,+)$ is a group generated
  by $U\subseteq G$,
  and suppose that $\tau\colon X\times G\to X$ is a transitive action
  of $G$ on~$X$.  Consider the environment $\XX=(X,U,f,h)$ where
  $f=\tau\rest (X\times U)$ and $h$ is pointed. 
  Let
  $E=\MSR_{\FF(A)}(\hat f^{-1}_{x_0}(E_h))$. Then
  $$\FF(A)/E\cong \XX.$$
\end{Cor}
\begin{proofV}
  By Corollary~\ref{cor:Main1} it is enough to show that $\XX$ is
  connected and minimally distinguishing. Since the action is
  transitive, $\XX$ is connected. Suppose $x_1*u=x_2*u=x_0$
  for some $x_0,x_1,x_2\in X$ and $u\in U$. Multiplying by the
  inverse of $u$ on the right this implies that $x_1=x_2$.
\end{proofV}

\section{Examples}

\begin{figure}
    \centering
    \begin{tabular}{cc}
        \begin{subfigure}[b]{0.4\textwidth}
            \centering
            \includegraphics[width=\textwidth]{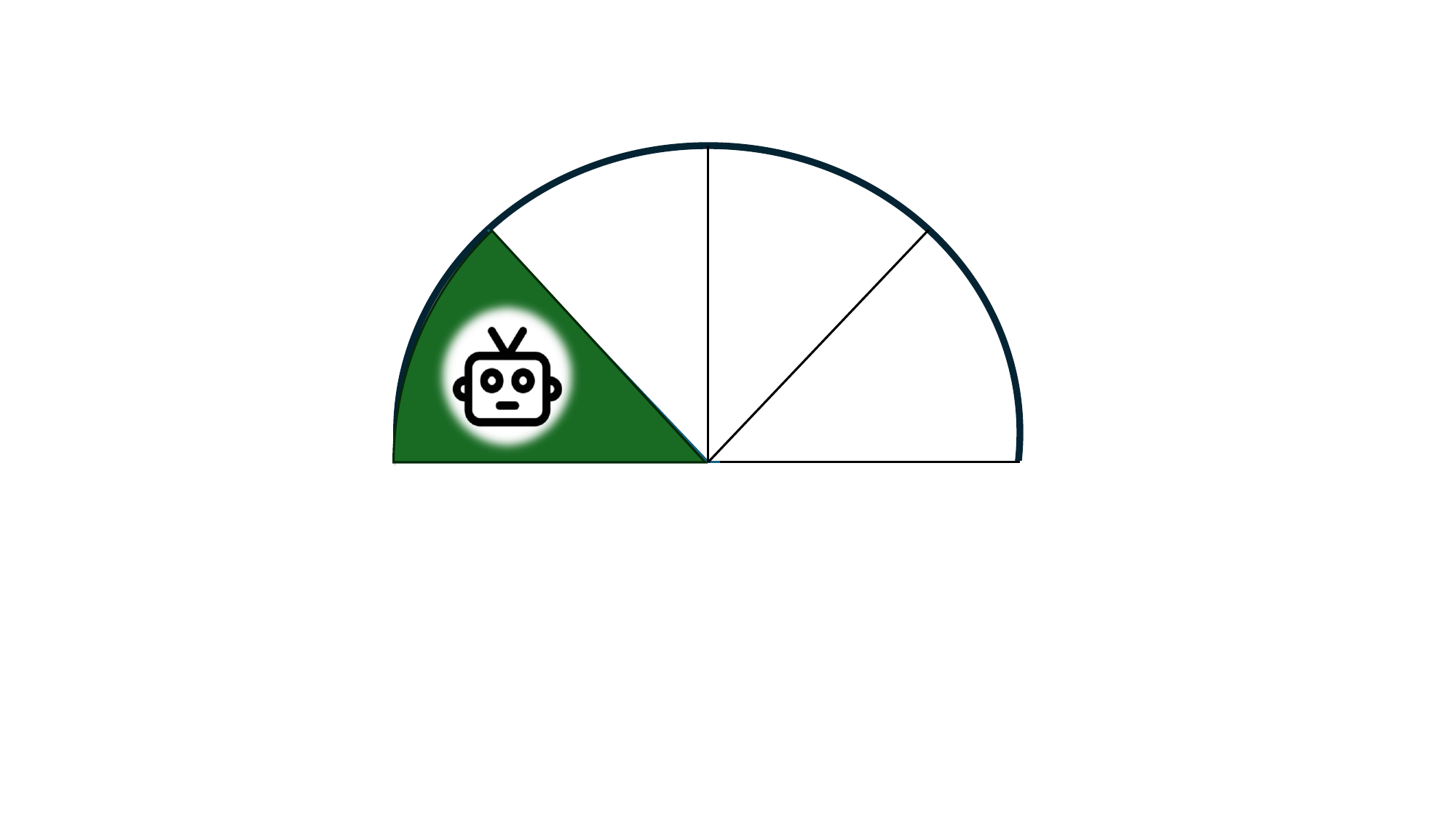}
            \caption{}
            \label{fig:subfig-a}
        \end{subfigure} &
        \begin{subfigure}[b]{0.4\textwidth}
            \centering
            \includegraphics[width=\textwidth]{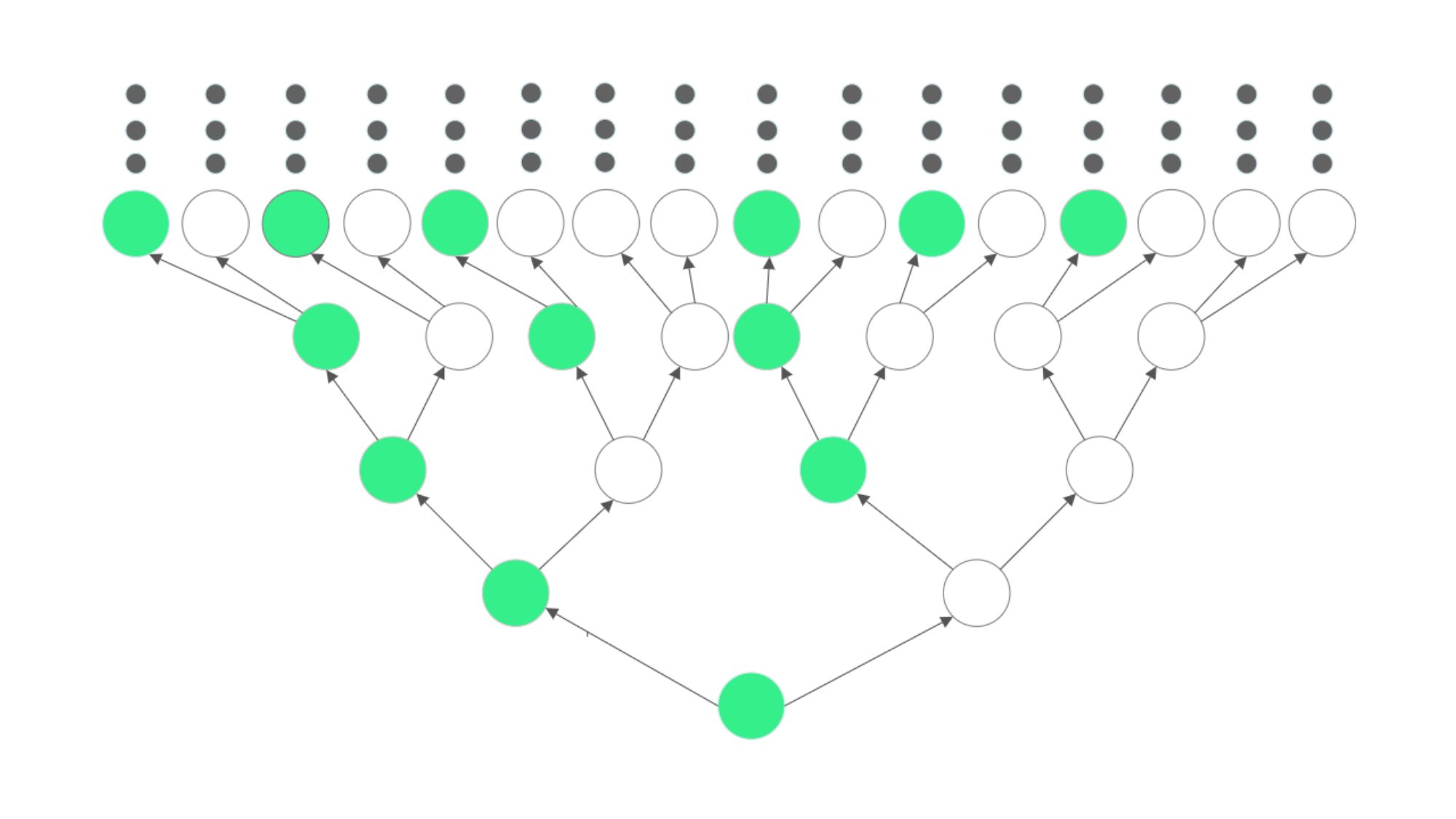}
            \caption{}
            \label{fig:subfig-b}
        \end{subfigure} \\
        \begin{subfigure}[b]{0.4\textwidth}
            \centering
            \includegraphics[width=\textwidth]{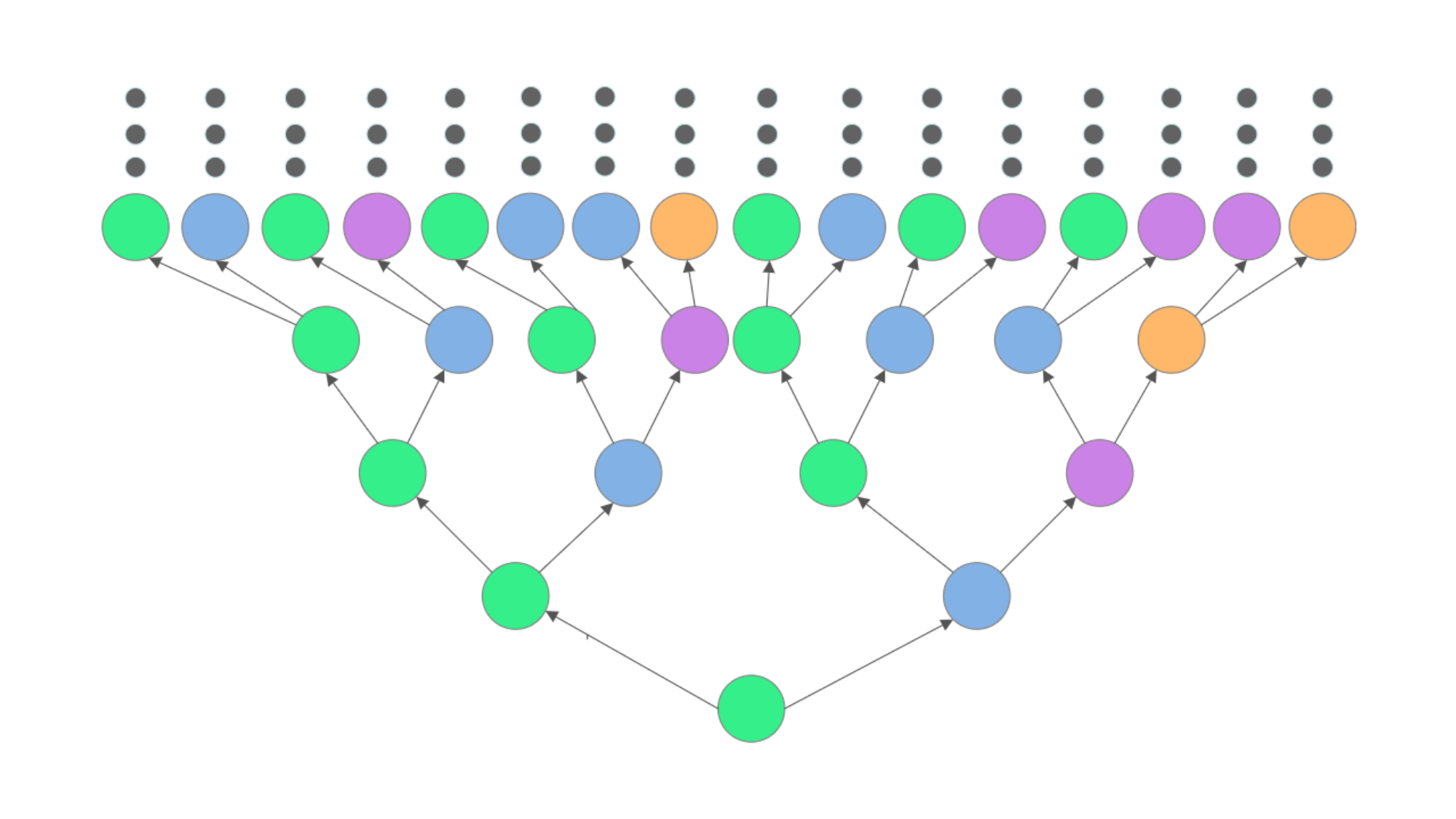}
            \caption{}
            \label{fig:subfig-c}
        \end{subfigure} &
        \begin{subfigure}[b]{0.4\textwidth}
            \centering
            \includegraphics[width=\textwidth]{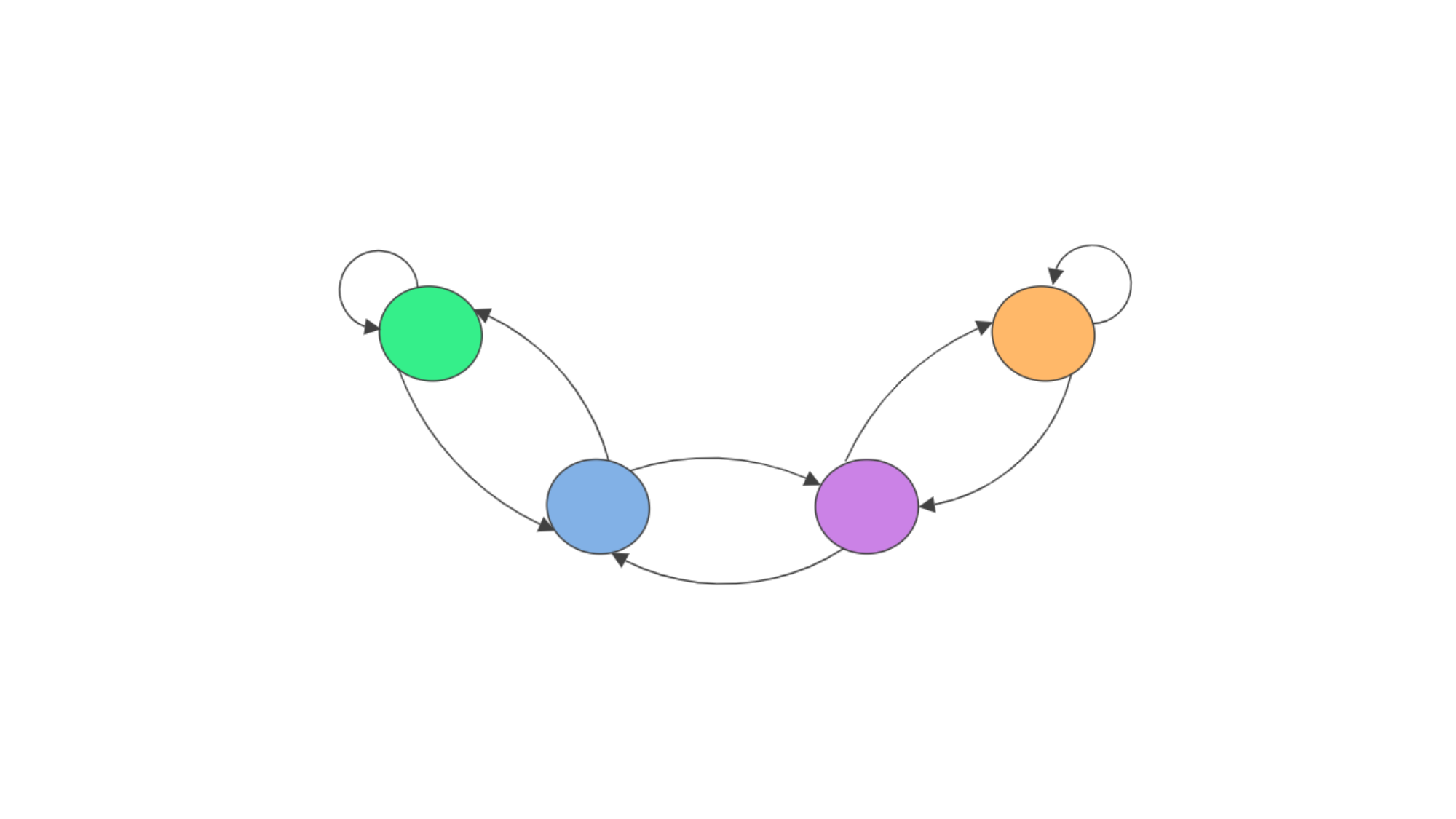}
            \caption{}
            \label{fig:subfig-d}
        \end{subfigure} 
    \end{tabular}
    \caption{The environment (a) has four states. Agent can move clockwise
    and counter-clockwise, but in states 1 and 4 nothing happens, if it tries
    to go left or right respectively. In (b) the binary tree of action observation
    sequences is depicted. In most nodes the sensor data is ``white'' but when 
    the agent is in the left-most state, the data is ``green''. In (c)
    we show the minimal sufficient refinement of (b), and in (d) we have taken
    the quotient of (c) with respect to the refinement. It turns out to be
    isomorphic to (a), as predicted by Corollary~\ref{cor:Main1}.
    \label{fig:Indistinguishable}}
\end{figure}

\begin{figure}
    \centering
    \begin{tabular}{cc}
        \begin{subfigure}[b]{0.4\textwidth}
            \centering
            \includegraphics[width=\textwidth]{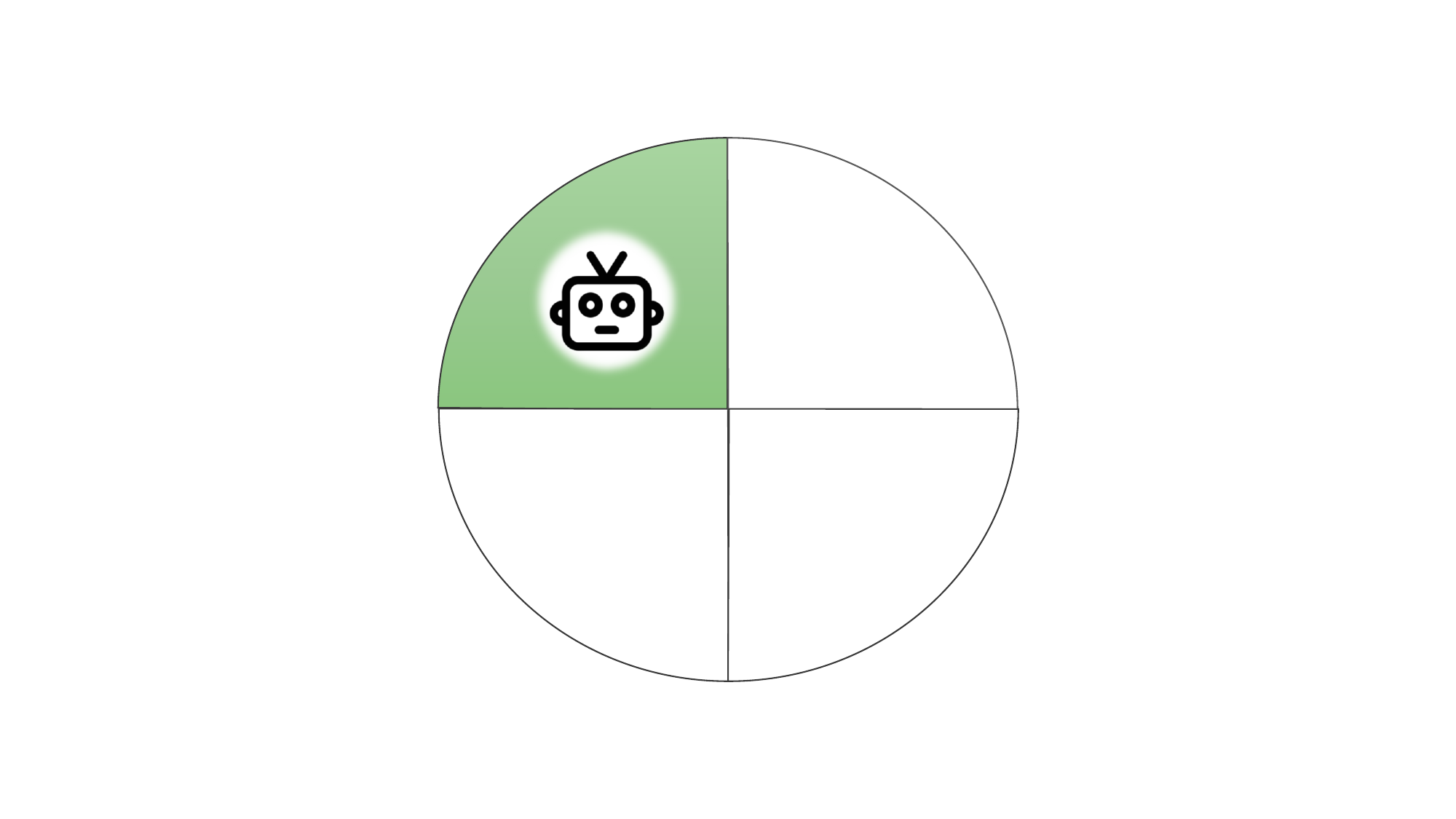}
            \caption{}
            \label{fig:subfig-a2}
        \end{subfigure} &
        \begin{subfigure}[b]{0.4\textwidth}
            \centering
            \includegraphics[width=\textwidth]{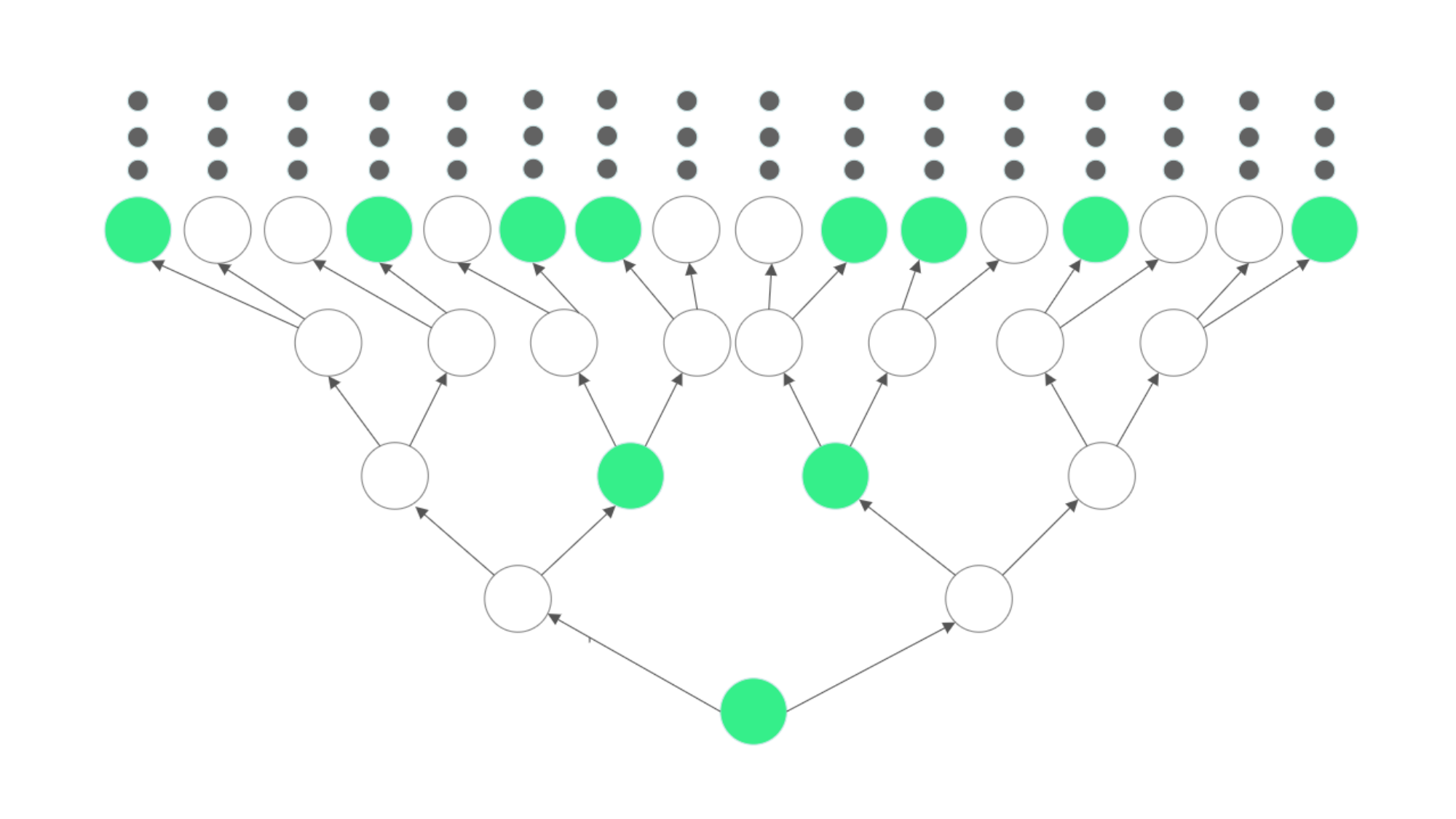}
            \caption{}
            \label{fig:subfig-b2}
        \end{subfigure} \\
        \begin{subfigure}[b]{0.4\textwidth}
            \centering
            \includegraphics[width=\textwidth]{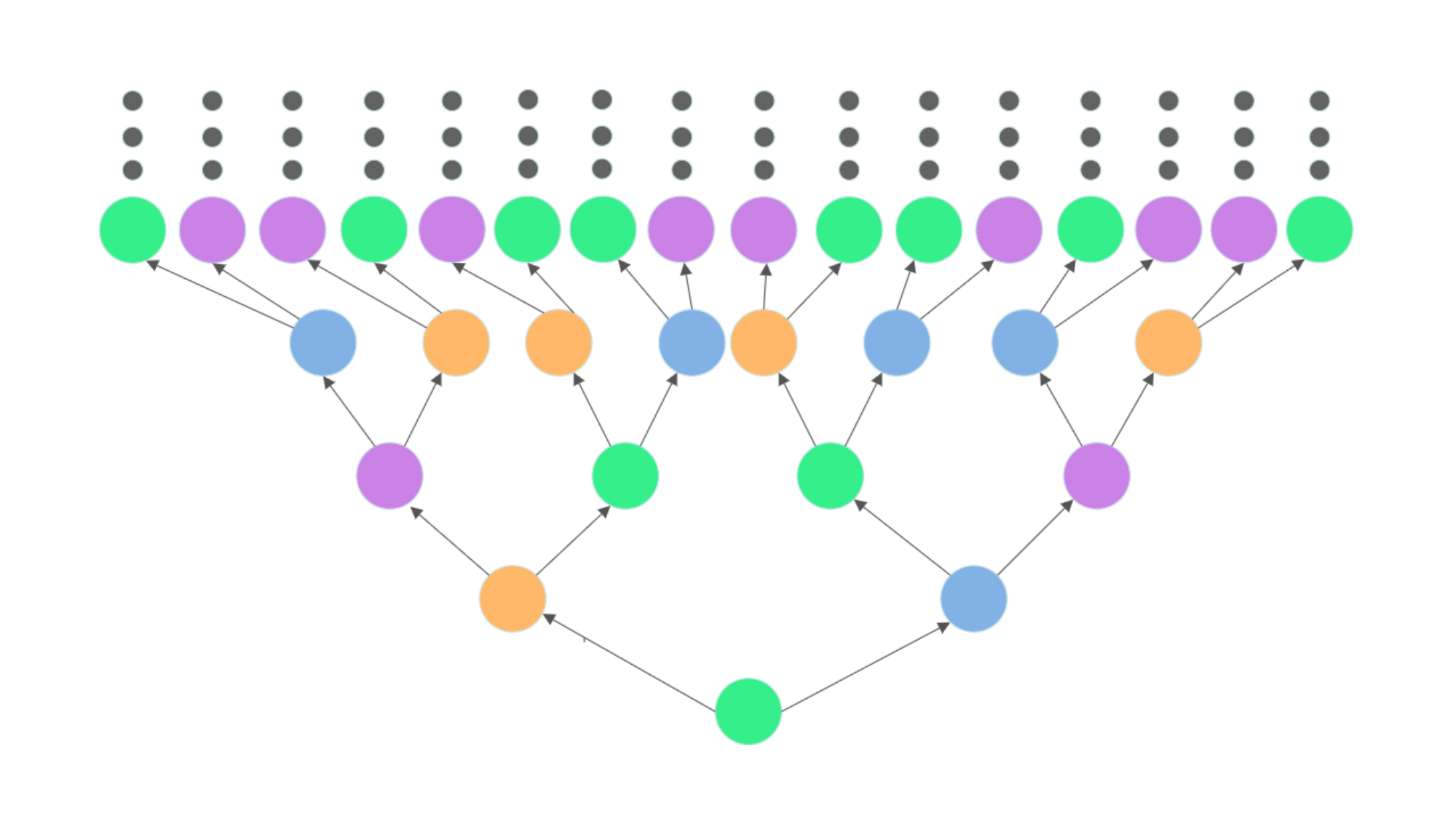}
            \caption{}
            \label{fig:subfig-c2}
        \end{subfigure} &
        \begin{subfigure}[b]{0.4\textwidth}
            \centering
            \includegraphics[width=\textwidth]{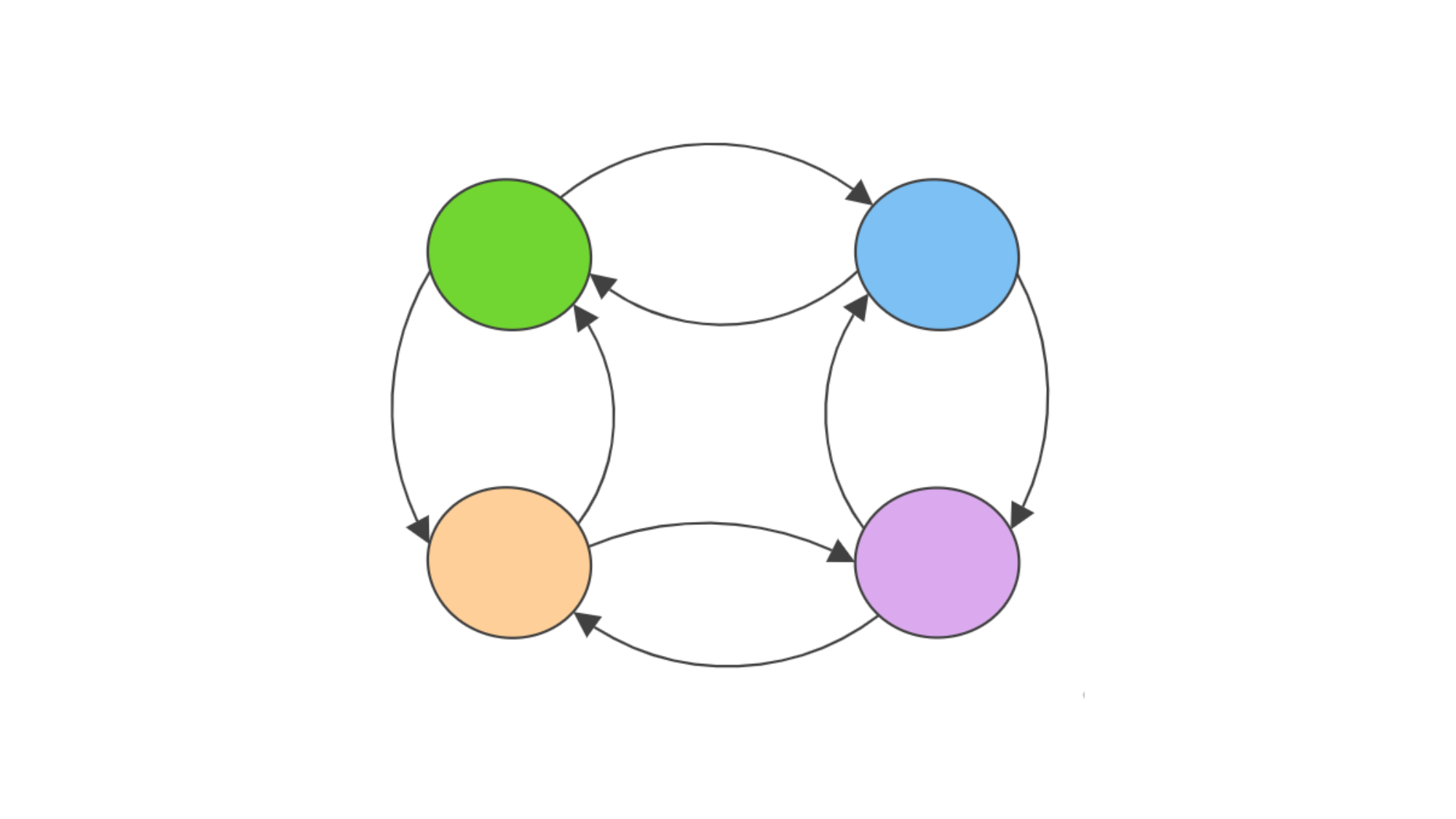}
            \caption{}
            \label{fig:subfig-d2}
        \end{subfigure} 
    \end{tabular}
    \caption{Same idea, as in Figure~\ref{fig:Indistinguishable}, but with a
    circular environment. In this case Corollary~\ref{cor:GroupAction} could
    be applied, with the group $\Z/4\Z$ acting on~$\XX$.}
    \label{fig:2}
\end{figure}

\begin{Example}\label{ex:Simple}
    Simple examples of finite external systems which can be ``learned''
    by taking minimal sufficient refinements over the 
    exploratory internal
    state space are depicted in Figures~\ref{fig:Indistinguishable}
    and \ref{fig:2}. It is easy to see that in both cases  the external state
    space is minimally distinguishing and the sensor mapping
    is pointed, so we can apply Corollary~\ref{cor:Main1}.
\end{Example}

In the final example below we show that very limited proprioceptive feedback
is enough for the emergent of internal structure isomorphic to the environment.

\begin{figure}
    \centering
    \includegraphics[width=0.6\textwidth]{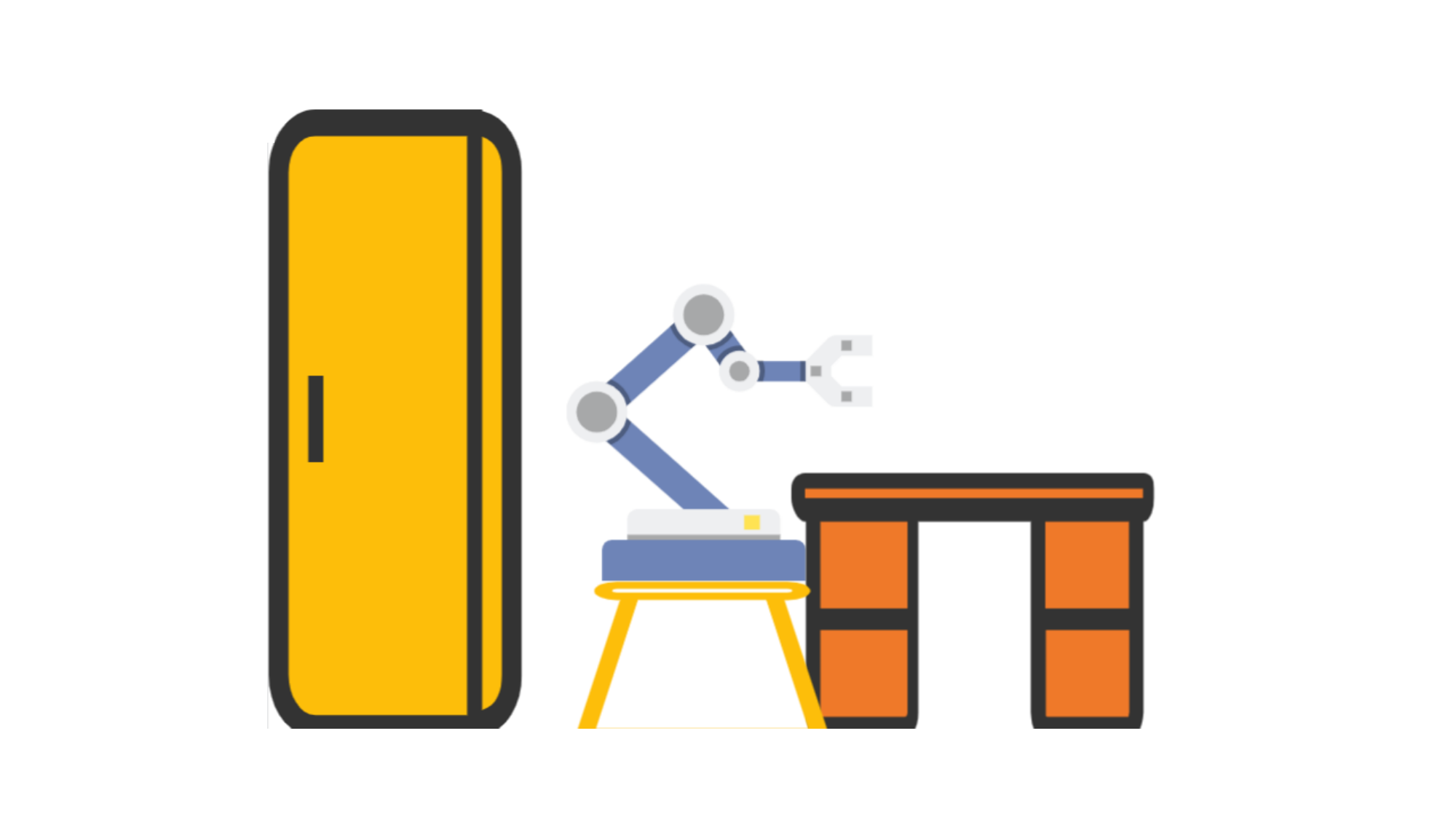}
    \caption{Robot arm with three joints and obstacles.}
    \label{fig:RA}
\end{figure}

\begin{Example}(Robot arm)\label{ex:RobotArm}
    In Figure~\ref{fig:RA} there is a robot arm with three joints.
    For the sake of applicability of our theory, assume that
    the state space is discrete. Each joint can turn
    by, say, $\pm 1^\circ$. If the arm is pushing against an obstacle
    while applying a rotation of any of the joints,
    then the external state stays the same as before. The configuration
    space is a subset of the (discrete) 3-torus. Assume that there is
    one (and possibly only one!) position of the arm where it receives a proprioceptive feedback. It could be the original position of the arm
    where it receives a ``click'' and no sensor feedback in any other position.
    Then this sensor mapping is pointed. It is also not hard to 
    see that the robot arm satisfies minimal distinguishability:
    Suppose the arm is  in some positions $x_1$ or $x_2$ and $u$
    $\pm 1^\circ$ rotation of one of the joints. If this rotation
    does not result in hitting an obstacle for either $x_1$ 
    or $x_2$, then $x_1\ne x_2$ implies $x_1*u\ne x_2*u$ because
    the rotation $u$ acts ``homeomorphically'' on the torus.
    If, however, say, $x_1$ faces an obstacle when rotating by $u$,
    then $x_1=x_1*u$, so we are done. Applying Corollary~\ref{cor:Main1}
    we see that if such robot arm succeeds in minimizing the surprise
    of ``when is the proprioceptive `click' feedback received'', then
    it will end up building an isomorphic copy of the external state space.
\end{Example}

\section{Discussion}
\label{sec:Discussion}

Our aim was to introduce a mathematical principle that the robot can internally evaluate, and if satisfied, ensures that the internal system becomes isomorphic or bisimulation equivalent to the environment. In doing so, we proposed a framework without pre-assigned meanings or assumed correlations between the internal and the external systems. Yet, a correlation emerges between them due to structural coupling. The proposed mathematical principle is sufficiency, which can be interpreted as the lack of surprise. In this way, 
our framework can be seen as an extension of the FEP framework to the combinatorial realm of finite or countable automata. We propose to explore the ideas of FEP further within this framework. 
One future direction is to explore the notion of Markov blankets in the context of discrete non-probabilistic systems. After all, the notion
of independence is more ubiquitous than that dictated by probability theory 
\mbox{\cite[Ch.2]{Baldwin}},~\cite{Paolini2015,Vaananen_2007}.

One drawback of our results is that both our
assumptions and conclusions are very strong. The assumption of sufficiency
is unrealistic, and the resulting similarity between the internal
and external models is too strong for most applications. Thus, another direction
for future research is to explore significantly weaker notions. For example, instead of minimizing surprise globally, the agent
may focus on minimizing surprise relative to its goals. 
We envision replacing minimal sufficient refinements by equivalences that retain only task-relevant information. 
This approach will guide the theory towards game-theoretic aspects, incorporating von Neumann's concepts of turn-taking and discrete strategies, as well as Nash's payoff functions and Aumann's epistemology works. 

We are also working on extending the framework to incorporate multiple sensing modalities
within one agent's brain, as well as  multiple agents, thereby uncovering new 
pathways in communication theory. 

\begin{credits}
\subsubsection{\ackname} 
Authors 1, 3, and 5 were supported by a European Research Council Advanced Grant (ERC AdG, ILLUSIVE: Foundations of Perception Engineering, 101020977).

\end{credits}
%
%
\bibliographystyle{splncs04}
\bibliography{references}

\end{document}